\pdfoutput=1

\documentclass[11pt]{article}
\usepackage{authblk}

\usepackage[]{EMNLP2023}

\usepackage{times}
\usepackage{latexsym}

\usepackage[T1]{fontenc}

\usepackage[utf8]{inputenc}

\usepackage{microtype}

\usepackage{inconsolata}

\usepackage{bibentry}
\usepackage{dialogue}
\usepackage{times}
\usepackage{latexsym}
\usepackage{mathrsfs}
\usepackage{amssymb}
\usepackage{amsmath}
\usepackage{url}
\usepackage{pifont}
\usepackage{graphicx}
\usepackage{color,xcolor}
\usepackage{braket}
\usepackage{bbold}
\usepackage[utf8]{inputenc}
\usepackage[toc,page]{appendix}
\usepackage{cleveref}
\usepackage{booktabs,graphicx,makecell,siunitx,eqparbox}
\usepackage[utf8]{inputenc}
\usepackage{color,soul}

\usepackage{enumitem}
\usepackage{float}
\usepackage{stmaryrd}

\makeatletter
\newcommand\avsuminner[2]{%
  {\sbox0{$\m@th#1\sum$}%
   \vphantom{\usebox0}%
   \ooalign{%
     \hidewidth
     \smash{\vrule height\dimexpr\ht0+1pt\relax depth\dimexpr\dp0+1pt\relax}%
     \hidewidth\cr
     $\m@th#1\sum$\cr
   }%
  }%
}
\makeatother

\title{Reformulating NLP tasks to Capture Longitudinal Manifestation of Language Disorders in People with Dementia.}

\author[1]{Dimitris Gkoumas}
\author[1,2,3]{Matthew Purver}
\author[1,2]{Maria Liakata}
\affil[1]{Queen Mary University of London, London, UK}
\affil[2]{The Alan Turing Institute, London, UK}
\affil[3]{Jožef Stefan Institute, Ljubljana, Slovenia}
\affil[ ]{\textit {\{d.gkoumas,m.purver,m.liakata\}@qmul.ac.uk}}

\begin{document}
\maketitle
\begin{abstract}
Dementia is associated with language disorders which impede communication. Here, we automatically learn linguistic disorder patterns by making use of a moderately-sized pre-trained language model and forcing it to focus on reformulated natural language processing (NLP) tasks and associated linguistic patterns. Our experiments show that NLP tasks that encapsulate contextual information and enhance the gradient signal with linguistic patterns benefit  performance. We then use the probability estimates from the best model to construct digital linguistic markers measuring the overall quality in communication and the intensity of a variety of language disorders. We investigate how the digital markers characterize dementia speech from a longitudinal perspective. We find that our proposed communication marker is able to robustly and reliably characterize the language of people with dementia, outperforming existing linguistic approaches; and shows external validity via significant correlation with clinical markers of behaviour. Finally, our proposed linguistic disorder markers provide useful insights into gradual language impairment associated with disease progression.

\end{abstract}

\section{Introduction}

Dementia is a neuro-degenerative disease affecting millions worldwide and is associated with cognitive decline, including language impairment~\cite{forbes2005detecting}. Language dysfunction may be difficult to detect in the early stages of dementia~\cite{nestor2004advances}; however, as the disease progresses, a gradual decline of semantic knowledge ensues, and eventually, all linguistic functions can be 
lost~\cite{tang2008assessment,klimova2015}. Recognizing language disorders as prodromal symptoms in people with dementia may help with earlier diagnosis and improve disease management.

Dementia can cause a variety of language deficits, such as: word-finding problems, a.k.a.\ \textit{anomia}~\cite{kempler2008language}; 
eloquent articulation lacking the expression of meaningful information, a.k.a.\ \textit{empty speech}~\cite{nicholas1985empty}; dropping speech, when the last few words in an utterance become barely audible 
a.k.a.\ \textit{trailing off speech}; or \textit{circumlocution} of words/concepts within an utterance~\cite{silagi2015naming}; interruptions in the smooth flow of speech, a.k.a.\ \textit{disfluency}~\cite{ferreira2004disfluencies}, characterized by repeated words, self-interruptions, and corrections of one's own speech, a.k.a. \textit{self-repair}~\cite{levelt1983};
\textit{agrammatism}, a syntactic disturbance, characterised by telegraphic speech, misuse of pronouns, or poor grammar~\cite{garre2018epidemiology}.

\begin{table*}[ht]
\centering 

\resizebox{\textwidth}{!}{%
\begin{tabular}{p{2.5cm}p{8cm}p{6.5cm}}
 \toprule
 
\textbf{Disorder} &  \textbf{Example Utterances} & \textbf{Symptoms/Manifestation in Language}   \\
 \hline
 Anomia & a) He's trying to get \textcolor{blue}{this} and he's gonna fall off of \textcolor{blue}{there} b) If that little girl \textcolor{blue}{don't} \textcolor{blue}{xxx}. c) The boy hasn't \textcolor{blue}{gotten down} to his \textcolor{blue}{fall} yet.  & a) Empty speech, b) trailing off speech, c) circumlocution in speech\\ \hline
 Disflueny &a) The wife is wiping a \textcolor{blue}{dish plate}. b) \textcolor{blue}{His his} sister's asking for one. c) Here's a \textcolor{blue}{sp} water spigot here .  &  a)  Word/phrase revision, b) word/phrase repetition, c) phonological fragment  \\ \hline
 Agrammatism & a) \textcolor{blue}{Water running down} from the sink. b) \textcolor{blue}{Her doing} the dishes. c) Three pieces \textcolor{blue}{of  to eat on}.  & a) Telegraphic speech, b) Misuse of pronouns c) poor grammar    \\ 
 
 \bottomrule
\end{tabular}
}
\caption{Language disorders associated with dementia and corresponding manifestation observed in the speech of subjects in the DementiaBank and ADReSS datasets. Words in blue denote linguistic disorder patterns. 
}\label{tbl:lan_dis_type}
\end{table*}

Table~\ref{tbl:lan_dis_type} 
provides the most common language disorders and associated manifestation (linguistic patterns) observed in the speech of subjects describing the Cookie Theft Picture 
(CTP, Appx. \ref{app:ctp}) in the DementiaBank~\cite{becker1994natural} and ADReSS~\cite{luz2020} datasets. Here we use state-of-the-art Natural Language Processing (NLP) to learn linguistic patterns indicative of language disorders in transcribed speech from people with dementia and healthy controls. We subsequently use the resulting language models to characterise the language of individuals with dementia. 

Early work in NLP for dementia relied on manual engineered features based on specific lexical, acoustic and syntactic features stemming from description tasks (such as CTP), to detect linguistic signs of cognitive decline~\cite{fraser2016linguistic,beltrami2018speech,yeung2021}. Recent work uses naive neural approaches to classify and analyse linguistic and acoustic characteristics so as to either predict cognitive scores or achieve binary classification of participants (Alzheimer’s Disease (AD) vs non-AD)~\cite{karlekar2018detecting,balagopalan2020bert,nasreen2021,rohanian2021}. However, such approaches tend to learn language discrimination across cohorts ignoring explicit information entailed in linguistic patterns within the language itself. This is because the optimization objective is to learn a unified label space and thus important linguistic patterns never have any gradient signal~\cite{tam2021improving}. Moreover previous work ignores the longitudinal aspect of language disorders. 
Here, we address these limitations and make the following contributions:
\begin{itemize}[noitemsep,topsep=0pt,parsep=0pt,partopsep=0pt,leftmargin=*]

\item We learn a variety of linguistic patterns characteristic of language disorders from transcribed  utterances by people with dementia and healthy controls. To achieve this we force a moderately-sized pre-trained LM, namely RoBERTa~\cite{liu2019roberta}, to focus on reformulated NLP tasks (Sec.~\ref{sec:finetune_tech}). To the best of our knowledge, ours is the first attempt to apply 
the recent successful NLP paradigm shift 
of reformulating  classification as text-to-text generation~\cite{tam2021improving,wang2021entailment,liu2023pre} in the context of dementia and mental health more broadly. We 
show that tasks encapsulating context and forcing the model to extract signal from the language itself 
benefit performance (Sec.~\ref{sec:quantitative}).
\item We introduce human interpretable digital linguistic markers to measure the quality of communication as well as the extent of a variety of language disorders in people with dementia. To construct the digital markers we leverage the model's probability estimates (Sec.~\ref{sec:task}).
\item We conduct a comprehensive longitudinal analysis to investigate how the linguistic communication marker characterizes individuals' speech. 
This shows significant discrimination across healthy controls, people with mild cognitive impairment (MCI), and people with AD (Sec.~\ref{sec:discrimination}).
\item We compare our proposed communication marker against existing 
approaches based on semantic similarity and word-level disfluency; 
ours shows better diagnostic performance 
(Sec.~\ref{sec:discrimination}). 
\item We evaluate the reliability of the communication marker against two clinical markers of behaviour widely used for assessing dementia and show significant correlation 
(Sec.~\ref{sec:reliability}).
\item We show that the proposed linguistic disorder markers  provide useful insights into the gradual language impairment associated with disease progression (Sec.~\ref{sec:disorder_markers}).

\end{itemize}
\section{Related Work}
\subsection{NLP for Dementia}
 Early NLP work for dementia detection analysed manually aspects of language such as lexical, grammatical, and semantic features \cite{ahmed2013connected,orimaye2017predicting,kave2018severity}, para-linguistic features \cite{gayraud2011syntactic,lopez2013selection,pistono2019happens}, and interactional patterns in conversations~\cite{elsey2015towards}.
 
 Recent work has made use of manually engineered features \cite{luz2020,luz2021detecting,nasreen2021detecting}, disfluency features \cite{nasreen2021,rohanian2021}, or acoustic embeddings \cite{yuan2020disfluencies,shor2020towards,pan2021using,zhu2021wavbert}. All such previous work has focused on differentiating across cohorts, without considering language changes over time or the importance of emergent linguistic patterns. 
 Recent work examined the longitudinal changes relies only on speech from public figures~\citep{petti2023}.

\subsection{Language Models}
Language models, the prevalent technology within NLP, are usually trained with the Cloze objective where part of the context in a text is removed, and the model is tasked with predicting the missing text~\cite{taylor1953cloze}. Masked language modeling (MLM) is a Cloze-based denoising objective that has been widely used in pre-training language model~\cite{yang2022learning}. Several works have reformulated learning tasks as cloze questions to re-purpose pre-trained language models ~\cite{schick2020exploiting,liu2023pre}. Other works have exploited task descriptions (prompts) and annotated examples with demonstrations to enable few-shot learning for downstream  tasks~\cite{gao2020making,wang2021entailment}. Such approaches have become an important research field as they overcome the challenge of expensive data annotation~\cite{li2022pre}. However, finding ways to reformulate tasks as cloze questions that make the best use of knowledge stored in language models can be difficult~\cite{schick2020s}. Here we follow  the task reformulation paradigm to force a model to learn linguistic patterns of language disorders.
\section{Problem Setup}

\subsection{Task Definition}
\label{sec:task}
Our task is that of learning linguistic patterns of language disorders framed as a multi-class classification problem. This involves fine-tuning a pre-trained language model \( \mathcal{L} \) on a collection of \( \mathcal{N} \) transcribed speech utterances $\{u_i\}_{i=1}^N$ from people with dementia and healthy controls elicited by the CTP description task. Here, an utterance is an unbroken chain of spoken language, so it may map to a sentence, part of a sentence or include multiple sentences. Each utterance is mapped to a single label $y_{i} \in$ \( \mathcal{Y}=\{anomia, disfluency, agrammatism, fluent\}$~\footnote{The label \textit{fluent} indicates an utterance does not exhibit any of the linguistic disorder patterns. Only $165/4037$ samples in the DementiaBank and ADReSS corpora have two labels, so we frame it as a single-label multi-class task.} and the goal is to predict the corresponding label. 
During fine-tuning emphasis is placed on strategies for reformulating the classification task into different NLP tasks.

For evaluation purposes we construct digital markers using the probability estimates of the model, to capture the overall quality in communication and the intensity of each of the language disorders. 
For the communication marker, we first extract the model's output probability estimate of an utterance to be fluent, i.e., $p(y_i^\mathcal{L} \; | \; y_{i}=fluent)$, and then obtain averaged probabilities over the entire session (description of the CTP). To investigate the discriminating ability of the communication marker across cohorts, we calculate average and longitudinal changes in the marker. To assess its reliability, we investigate the association between changes in this marker compared to two widely used clinical behavioural markers over time (Sec.~\ref{sec:metrics}) . 
We similarly construct anomia, disfluency, and agrammatism markers (see Appx.~\ref{app:train}), 
and compare their changes
across cohorts as above. 

\subsection{Data}
\label{sec:data}
We conduct experiments and train models on transcribed speech from two datasets, namely ADReSS~\cite{luz2020} and DementiaBank~\cite{becker1994natural}. They both contain transcribed speech of people with dementia and healthy controls describing the 
Cookie Theft Picture (Appx.~\ref{app:ctp}). ADReSS includes a single speech sample per participant while DementiaBank contains longitudinal speech, up to five times per person (see Appx.~\ref{app:pitt} for a detailed description of the datasets). For training models, we use data from ADReSS and also transcripts from subjects who contributed up to two descriptions in DementiaBank. Table~\ref{tbl:data_train} provides an overview of the datasets. Utterance annotations are based on the paralinguistic information available in transcribed scripts using the 
CHAT protocol~\cite{macwhinney2017tools}. For details about the coding scheme please refer to Appx.~\ref{app:chat}. During  pre-prossessing, we remove the paralinguistic information and discard the carers' utterances as well as patients' non-descriptive utterances. We split the data into training (80\%), validation (10\%) and testing (10\%) keeping same class proportions across the splits. The split is done in a way that only utterances are unseen in the test set. Hence,  users might be seen in the test set. 

\begin{table}[htbp]
\centering 
\resizebox{\columnwidth}{!}{%
\begin{tabular}{lcccccc}
 \toprule
\textbf{Cohort} & \textbf{\# Sub.} & \textbf{\# Ses.} & \textbf{\# Flt.} & \textbf{\# Ano.} & \textbf{\# Dis.} &\textbf{\# Agr.}    \\
 \hline
 Healthy & 107 & 136 & 908 & 9 & 246 & 195 \\ \hline
 Dementia & 224 & 277 & 1337 & 203 & 734 & 405\\ 
 \bottomrule
\end{tabular}
}
\caption{Statistical overview  of utterance-level annotations in ADReSS and DementiaBank used for training. Abbreviations: Sub.=Subjects, Ses.=Sessions, Flt.=Fluent, Ano.=Anomia, Dis.=Disflunecy, Agr.=Agrammatism.} 
\label{tbl:data_train}
\end{table}

To conduct a longitudinal evaluation
we use a subset  
from DementiaBank 
of 
healthy controls and people with dementia who have 3, 4 and 5 sessions. The corresponding numbers for controls are 28/10/8 and for people with dementia 12/8/3.

\subsection{Fine-Tuning Strategies and NLP tasks}


\label{sec:finetune_tech}

We take a moderately sized pre-trained language model (PLM) \(\mathcal{L}\) = RoBERTa~\cite{liu2019roberta} and fine-tune it according to different strategies.

\paragraph{Standard Fine-tuning ($\mathcal{L}_{standard-finetune}$):} \label{par:fine-tune} Given the PLM  \( \mathcal{L} \), we first convert an utterance $u$ into a sequence of tokens 
$\overline{u}$ = $[CLS] \: t_1 \: t_2 \dots t_n [SEP]$ where $t_1 \dots t_n$ are the tokens in utterance $u$ \footnote{$\overline{u}$ is defined in the same way for all the tasks.}. The model takes $\overline{u}$ and maps the original utterance to a sequence of logits $\mathcal{L}(\overline{u}) \in \mathbb{R}^{|\mathcal{Y}|}$. At prediction time, softmax is applied for multiclass classification. We fine-tune the model with cross-entropy loss as follows:
\begin{equation}
\label{eq:ce}
    Loss_{} =  CE(p(y^\mathcal{L}|\overline{u}), y)
\end{equation}
where $p(y^\mathcal{L}|\overline{u})$ is softmax over $y$ calculated as:  
\begin{equation}
    p(y^\mathcal{L}|\overline{u}) = \frac{exp(\llbracket \mathcal{L}(\overline{u}) \rrbracket)_y }{\sum_{y' \in \mathcal{Y}} exp(\llbracket \mathcal{L}(\overline{u}) \rrbracket)_y'}
\end{equation}

\paragraph{Multitask Fine-tuning with MLM:} We fine-tune the PLM  \( \mathcal{L} \) with two objectives. The first one is the masked language model (MLM) objective to understand particular linguistic patterns in the domain. We first convert an utterance $u$ to a sequence of tokens $\overline{u}$ as above and then dynamically~\footnote{Different tokens are randomly masked in each epoch.} mask 15\% of tokens within the utterance~\cite{devlin2018bert}. For a given utterance $u$ (e.g., A mother is wiping a dish), the model receives a MLM input as  
\begin{equation*}
 [CLS] \; A \; mother \; \textbf{[MASK]} \; wiping \; a \; dish \; [SEP]  
\end{equation*}
and maps $[MASK]$ to a sequence of logits $\mathcal{L}(\overline{u}) \in \mathbb{R}^{|\mathcal{V}|}$, where $\mathcal{V}$ is the vocabulary of $\mathcal{L}$. The training process thus becomes a high-dimensional multi-class classification problem of predicting the original token corresponding to $[MASK]$ with cross-entropy loss (Eq.~\ref{eq:ce}). The second objective is to predict the class label $y_i \in \mathcal{Y}$ corresponding to an utterance $u$. 
(See~\ref{par:fine-tune}). 
We experiment with two variants: a) separate multitask learning, where each task is learned independently ($\mathcal{L}_{multitask-MLM-separately}$). We first fine-tuning the model on the MLM objective and then resuming fine-tuning for the second objective; b) jointly learning both objectives ($\mathcal{L}_{multitask-MLM-joint}$). The combined loss is a linear weighted sum of loss functions of the two objectives. The assignment of weights is an open research question. Here, we set the weights empirically, 
based on the minimum loss function values when fine-tuning the model on the two objectives separately (See Appx.~\ref{app:train}).

\paragraph{Entailment-based Fine-tuning ($\mathcal{L}_{entailment}$):} The goal here is to map the relationship between an utterance $u$ and the corresponding language disorder label to a relationship space by reformulating multi-class classification as an entailmenttask~\cite{wang2021entailment}, a.k.a. natural language inference (NLI). Here, a language disorder definition is assumed to entail utterance $u$ if the definition can be logically derived from utterance $u$, (e.g., for the utterance \textit{``His his sister's asking for one''} entails \textit{``Word repetition or revision''}).

Given an instance $(u,y)$, we construct a set of tuples $\{(u, p_j)\}_{j=1}^{|\mathcal{Y}|}$ for each class $y \in \mathcal{Y}$ where $\{p_j\}$ is a set of label definitions, including~\footnote{The label definitions were created on the basis of the CHAT protocol guidelines and manual analysis of the data } $\{$\textit{Talking around words/empty speech/incomplete speech, Word repetition or revision, Agrammatism or paragrammatism in speech, Fluent speech}$\}$. For each utterance, the model $\mathcal{L}$ receives a set of $|\mathcal{Y}|$ tuples~\footnote{This approach requires $|\mathcal{Y}|$ forward passes during inference time.} in the form: 
\begin{equation*}
 [CLS] \; u \; [SEP] \;  p_j \;  [SEP],
\end{equation*}
and outputs a sequence of logits $\mathcal{L}(u,p_j) \in \mathbb{R}^{|\mathcal{Y}| \times |\mathcal{E}|}$, where $\mathcal{E}=\{$\textit{entails, does not entail}$\}$. At inference time, we extract the probability of $p(entails | (u, p_j))$ for each class in $\mathcal{Y}$ and apply $argmax$ across the extracted probabilities. We fine-tune the model with cross-entropy loss.

\paragraph{Prompt-based Learning:} Here the  PLM  \( \mathcal{L} \) is tasked with "auto-completing" natural language prompts~\cite{liu2023pre}. In particular, for each utterance $u$ 
let  \( \mathcal{T} \)$(u)$ be a MLM input with one $[MASK]$ token. Let $\mathcal{M}:\mathcal{Y} \to \mathcal{V}^{|\mathcal{Y}|}$ be a one-to-one mapping from the task label space $\mathcal{Y}$ to individual words in the vocabulary $\mathcal{V }$ of $\mathcal{L}$. The model $\mathcal{L}$ receives a template $\mathcal{T}(u)$ and maps the $[MASK]$ token to a sequence of logits $\mathcal{L}(\mathcal{T}(u)) \in \mathbb{R}^{|\mathcal{V}|}$.  We cast the problem of predicting the probability of $y \in \mathcal{Y}$ as a MLM task: 
\begin{multline}
    p(y\;|\;u) = p([MASK] = \mathcal{M}(y) \; | \; \mathcal{T}(u)).
\end{multline}
For a set of instances $\{u, y\}$, $\mathcal{L}$ is fine-tuned to minimize the cross-entropy loss. 

\noindent We experiment with the following variants:

\begin{itemize}[noitemsep,topsep=0pt,parsep=0pt,partopsep=0pt,leftmargin=*]
\item \textbf{Standard Prompt-based} ($\mathcal{L}_{standart-prompt}$): 
Here the MLM consists of an utterance $u$ and a task-specific prompt as follows:
\begin{equation}
\label{eq:template}
\mathcal{T}(u) = [CLS] \: u \: . \: \underline {It \: is \:\textbf{[MASK]} \: .} \: [SEP]
\end{equation},
where the underlined text is the task specific template and $[MASK] \in \mathcal{M}(y)$. 
\item \textbf{Prompt-based with Demonstration Examples} ($\mathcal{L}_{prompt-demonstrations}$): We adopt the idea of incorporating demonstrations as additional context~\cite{gao2020making}. For each utterance $u$, we randomly sample one example  $(u,\mathcal{M}(y_i))_{i=1}^{|\mathcal{Y}|}$ from each class $y \in \mathcal{Y}$ and combine the original utterance and examples to create templates according to Eq.~\ref{eq:template}. For the random samples, we replace the $[MASK]$ token with $\mathcal{M}(y_i)$. The model $\mathcal{L}$ receives as input a combination of the templates: 
\begin{equation}
\label{eq:contextual}
\mathcal{T}(u) \: \oplus \: \mathcal{T}(u,\mathcal{M}(y_1))  \: \oplus \:  ...  \: \oplus \: \mathcal{T}(u,\mathcal{M}(y_i))
\end{equation}
where $\oplus$ denotes concatenation. Given a contextual utterance in the form of Eq~\ref{eq:contextual}, the task involves predicting the $[MASK]$ token in the original utterance. At test time we sample demonstration examples from the training subset.

\item \textbf{Prompt-based with Inverse Learning Objective ($\mathcal{L}_{prompt-inverse}$):} The standard prompt-based objective encapsulates the question \textit{``Given the input what is the right label''}. Here, we inverse the question, \textit{``Given the answer label, what is the correct content''}. The model $\mathcal{L}$ is trained on the objective of predicting the input given the label. Formally, an utterance $u$ is reformulated through $\mathcal{T}$ according to Eq.~\ref{eq:template}. Then, we replace the $[MASK]$ token in Eq.~\ref{eq:template} with the original class token $\mathcal{M}(y)$ and apply a 50\% random masking  across the utterance's tokens. Thus, we force the model to predict the tokens in the context of the original label $\mathcal{M}(y)$. The model outputs for each of the $[MASK]$ tokens a sequence of logits $\mathcal{L}(u) \in \mathbb{R}^{|\mathcal{V}|}$, where $\mathcal{V}$ is the vocabulary of $\mathcal{L}$. Similarly to the MLM objective, we apply cross-entropy loss to predict the masked tokens. At test time, we give the model the correct and incorrect labels $\mathcal{M}(y)$ and reform the utterance $u$ through $\mathcal{T}$. Out of $|\mathcal{Y}|$ combinations, we choose the one with minimum loss. 

\paragraph{Random Rate:} Finally we include weighted guessing as a baseline  classifier where accuracy is guessed at the weighted percentages of classes.
\\ \\
For the experimental settings when training RoBERTa across different NLP tasks, we refer readers to~\ref{app:train}.

\end{itemize}

\subsection{Evaluation Metrics}
\label{sec:metrics}
To evaluate the success of different NLP task reformulation strategies in capturing the different language disorders, we report per class accuracy and $F1$. We also calculate the macro-averaged accuracy and $F1$ score. We chose macro-averaged scores since we are interested in minority classes, such as anomia, important in charecterizing the communication ability of people with dementia. 

We evaluate the digital linguistic markers defined in Sec.~\ref{sec:task} against two widely used clinical behavioural markers, namely, the Mini-Mental State Examination (MMSE), and the Clinical Dementia Rating (CDR) scale~\cite{morris1997clinical}. The higher the MMSE score, the higher the cognitive function. In contrast, the higher the CDR, the lower the cognitive function. For a detailed description of the behavioural markers see Appx.~\ref{app:biomarkers}. 
\section{Experimental Results}

\subsection{Quantitative Results}
\label{sec:quantitative}
The motivation of this work is to learn various linguistic disorder patterns forcing the models to explicitly leverage information from the language itself rather than learning a unified space where important linguistic patterns never have any gradient signal over optimization. Therefore, standard-fine tuning, rather than random guessing, is our fundamental baseline since it does not take into account explicit linguistic patterns over the optimization. To this end, we report the deviation of the other fine-tuning strategies (numbers in parentheses, Table~\ref{tbl:experiments}) from the performance of standard fine-tuning.  

Table~\ref{tbl:experiments} summarizes the experimental results for NLP task reformulation for identifying language disorder patterns in transcribed speech from the DementiaBank and ADReSS datasets.  All fine-tuning and learning strategies yielded  significantly better performance than random weighted guessing. However, class imbalance has caused bias towards the majority class (i.e., fluent speech), leading to underperformance for the minority class (i.e., anomia). We also noticed a trade-off in performance between the majority and minority classes. We suppose this is because speech with anomia is still fluent and prosodically correct but overall meaningless.

\begin{table*}[htbp]
\centering 
\resizebox{\textwidth}{!}{%
\begin{tabular}{lcccccccccc}
 \toprule
 \multicolumn{1}{c}{} &
 \multicolumn{2}{c}{Fluent} &
 \multicolumn{2}{c}{Anomia} &
 \multicolumn{2}{c}{Disfluency} & 
 \multicolumn{2}{c}{Agrammatism} &
  \multicolumn{2}{c}{Macro}
 \\
\cmidrule(r){2-3} \cmidrule(r){4-5} \cmidrule(r){6-7} \cmidrule(r){8-9} \cmidrule(r){10-11}
\textbf{} & \textbf{Acc.} & \textbf{F$_1$} & \textbf{Acc.} & \textbf{F$_1$}  & \textbf{Acc.}  & \textbf{F$_1$} & \textbf{Acc.}  & \textbf{F$_1$} & \textbf{Acc.}  & \textbf{F$_1$}    \\
 \hline
 Random Rate & 30.8 & - & 0.2 & - & 5.7  & -  & 2.1 & - & 1.2 ($\downarrow$ 63.9) & -\\ \hline \hline
$\mathcal{L}_{standard-finetune}$ & \textbf{96.8} & 94.2 & 20.8 & 31.2 & 86.5  & 78.0 & 56.5 & 67.2 & 65.1 &67.5\\ \hline
$\mathcal{L}_{multitask-MLM-separately}$ &94.8  & 91.7 & 29.2 & 37.8 & 85.6  & 78.8& 50.7 & 61.9  & 65.1 ($\leftrightarrow$0.0) & 67.6 ($\uparrow$ 0.1)\\ \hline
$\mathcal{L}_{multitask-MLM-joint}$ & 93.7  & 92.0 & \textbf{45.8} & \textbf{48.9} & 74.8  & 71.6 & 55.1  & 62.3  & 67.3 ($\uparrow$ 2.2)  & 68.7 ($\uparrow$ 1.2)\\ \hline
$\mathcal{L}_{entailment}$ & 94.7 & 94.7 & 30.2 & 41.0 & \textbf{88.9}  & 76.3& 59.0 & 66.0 & 68.3 ($\uparrow$ 3.2)   & 70.3 ($\uparrow$ 2.8)\\ \hline \hline
$\mathcal{L}_{standard-prompt}$ & 96.4 & 93.1 & 29.2 & 41.2&  86.5  & 79.3 & 55.1  & 66.7  & 66.8 ($\uparrow$ 1.7) & 70.1 ($\uparrow$ 2.6)\\ \hline
$\mathcal{L}_{prompt-demonstrations}$ & 96.6 & \textbf{95.2} & 27.0 & 37.4 & 87.5  & \textbf{81.0} & \textbf{66.2} & \textbf{71.9} & \textbf{69.9} ($\uparrow$ 4.8) & \textbf{72.2} ($\uparrow$ 4.7)\\ \hline
$\mathcal{L}_{prompt-inverse}$ & 48.0 & 54.6 & 33.3 & 13.6 & 18.9  & 24.4 & 46.4 &  35.8& 36.7 ($\downarrow$ 28.4) & 25.7 ($\downarrow$ 41.8)\\  
 \bottomrule
\end{tabular}
}
\caption{Performance of models resulting from reformulated NLP tasks using RoBERTa for identifying language disorder patterns in transcribed speech from  the DementiaBank and ADReSS datasets. Numbers in bold indicate best performance. Numbers in parentheses denote deviation from the performance of standard fine-tuning.}
\label{tbl:experiments}
\end{table*}

Both multitask with MLM and inverse prompt-based learning tasks were trained with the objective of forcing the model to obtain signal from linguistic patterns associated with a unified label space. Joint multitask learning with MLM is robust with respect to the minority class. In particular, it achieves the best accuracy and $f_1$ scores for the anomia class compared to all other settings. On the other hand, prompt-based with inverse learning objective underperforms all other approaches. We assume this is because the latter does not have a gradient signal from the labels during optimization. 
This setting may be more appropriate when masking is targeted rather than random.
However, this would require word-level annotations which are not currently available in these datasets.

Tasks incorporating context in the form of additional information exhibit superior performance over tasks learning a unified space without context.  In particular, entailment-based fine-tuning which includes label descriptions achieves an increased macro accuracy of 68.3\% compared to 65.1\% for standard fine-tuning. Similarly, prompt-based learning with demonstrations incorporating examples from each class yields an increased macro accuracy of 69.9\% compared to 66.8\% for standard prompt-based learning.   

Overall, the experiments show that tasks which include context in the form of additional information and force the model to obtain signal from linguistic patterns yield better  performance. In particular, prompt-based learning with demonstrations, which meets both of the above characteristics, achieves an increased macro accuracy of 69.9\%,  compared to 65.1\% for standard fine-tuning trained with an objective that ignores patterns from the language itself during the optimization process.

\subsection{Longitudinal Discrimination Ability}
\label{sec:discrimination}

Using the probability estimates of RoBERTa trained on prompt-based learning with demonstration examples to recognise linguistic disorders (which yielded the highest macro-F1), we have created a digital communication marker and language disorder markers (See Sec~\ref{sec:task} for more details). We analyze changes in the digital communication marker over time and across cohorts of people with AD, MCI and healthy controls. We calculate the average of the communication marker across the three cohorts (Table \ref{tbl:discr_marker}). The higher the score of the marker (1st column), the lower the impact of language disorders on communication. We observe that the marker decreases alongside disease severity. In particular, there is a significant difference in the marker's scores across the healthy, MCI, and AD cohorts.\footnote{\label{test}We use the nonparametric 
Mann-Whitney test to measure if the distribution of a variable is different in two groups.}

\begin{table*}[htbp]

\centering 

\resizebox{\textwidth}{!}{%
\begin{tabular}{llccccccccc}
 \toprule
 \multicolumn{1}{c}{} &
 \multicolumn{3}{c}{\textbf{Our communication marker}} &
 \multicolumn{3}{c}{\textbf{Semantic similarity marker}} &
 \multicolumn{3}{c}{\textbf{Word-level disfluency marker}}\\
\cmidrule(r){2-4} \cmidrule(r){5-7} \cmidrule(r){8-10}

\textbf{Cohort} & \textbf{Marker} & \textbf{$\Delta_{(end-start)}$} & \textbf{$\Delta_{(long)}$}  & \textbf{Marker} & \textbf{$\Delta_{(end-start)}$} & \textbf{$\Delta_{(long)}$} & \textbf{Marker} & \textbf{$\Delta_{(end-start)}$} & \textbf{$\Delta_{(long)}$}     \\
 \hline
 Healthy & \textbf{0.759 (0.164)} & +0.011 (0.162) &  +0.000 (0.106)& 0.296 (0.077) & +0.013 (0.107) & +0.009 (0.054) & 0.913 (0.064) & -0.005 (0.072) & -0.003 (0.030)\\ \hline
 MCI & \textbf{0.630 (0.224)}  & +0.010 (0.164)  & +0.010 (0.068)  &  0.299 (0.080) & -0.051 (0.077)  & -0.017 (0.031) & 0.879 (0.081) & +0.019 (0.100) & +0.005 (0.030)\\  \hline
 AD & \textbf{0.536 (0.201)} & \textbf{-0.229 (0.117)}  & \textbf{-0.120 (0.094)} & 0.270 (0.067) & +0.011 (0.890) & +0.001 (0.038) & 0.892 (0.075) & -0.026 (0.081) & -0.008 (0.038) \\ 
 \bottomrule
\end{tabular}
}
\caption{Comparison of our proposed digital linguistic communication marker versus baselines from semantic similarity and word-level 
Marker: Average of marker within a population. $\Delta_{(end-start)}$: Average change of the marker from the end to the beginning of the study. $\Delta_{(long)}$: Average change of the digital marker between adjacent individuals' sessions. Positive number implies improvement over time. Numbers in $()$ refer to corresponding standard deviations. Numbers in bold denote significant difference across cohorts.}
\label{tbl:discr_marker}
\end{table*}

We subsequently calculate changes in the communication marker from the end to the start of the study and across cohorts (i.e.,  $\Delta_{(end-onset)}$ in Table \ref{tbl:discr_marker}). There is a significant decrease for the AD group compared to the healthy and MCI cohorts ($p < 0.05$) $^{\ref{test}}$. There was no significant change in linguistic ability for the MCI and healthy cohorts: 
for 
controls, there is presumably no cognitive decline; 
for the MCI group, 
changes in linguistic function are likely trivial~\cite{nestor2004advances}.

We also calculate changes in the communication marker between adjacent sessions over time and then aggregated them per individual. In Table~\ref{tbl:discr_marker}, we report the average change across cohorts, i.e., $\Delta_{(long)}$. We obtain similar results as the ones from the end to the start of the study.

We compare the discrimination ability of our communication marker against two baseline markers based on semantic similarity and word-level disfluency. For a baseline developed on semantic similarity, we use the Incoherence Model \cite{iter2018automatic}, which scores adjacent pairs of utterances based on the cosine similarities of their sentence embeddings \cite{reimers-2019-sentence-bert}. The higher the score, the better the thematic consistency within a session (CTP description). We note that the thematic consistency is higher for the MCI cohort compared to the healthy controls. However, there is no substantial difference across cohorts (see Table~\ref{tbl:discr_marker}, Semantic similarity marker). We observe similar results when analysing the semantic marker's longitudinal discrimination ability. For word-level disfluency, we use a pre-trained transformer model for word-by-word disfluency detection in the form of reparandum-interregnum-repair~\cite{rohanian2021best}. To construct the baseline marker, we use the normalized probability estimates of words within an utterance to be fluent and then average the scores obtained over a session (CTP description). The higher the score, the less the occurrence of disfluent patterns in speech. We obtain results similar to the ones from the semantic similarity marker. In particular, the score is higher for people with AD compared to those with MCI. However, there is no significant difference across cohorts. 


Overall, our proposed communication marker is robust and reliable in discriminating between people with dementia, MCI and healthy controls, identifying changes in linguistic ability over time and does so better than existing approaches.

\subsection{Communication marker Reliability}
\label{sec:reliability}

We investigate the reliability of the digital communication marker by associating longitudinal changes in the marker with two widely used behavioural measures collected over the study. We consider individuals across different cohorts with at least three sessions each (for the description of the evaluation dataset, see Sec.~\ref{sec:data}).

\begin{figure}[hbt!]
\centering
\includegraphics[width=.50\textwidth]{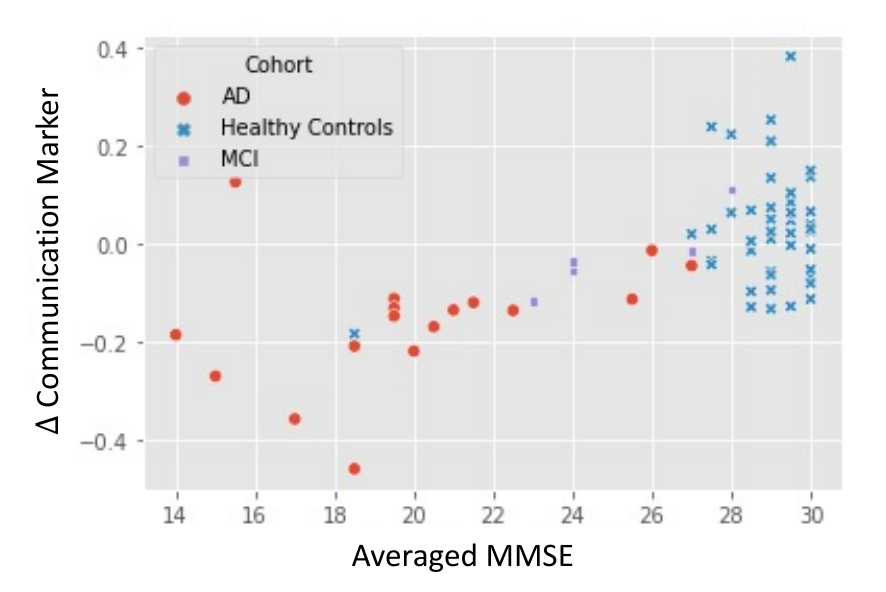}
\caption{Association between average 
longitudinal change in 
communication marker and average Mini-Mental State Examination (MMSE) scores, across 
cohorts. } 
\label{fig:mmse}
\end{figure}

We first investigate the association between longitudinal changes in the digital communication marker and the Mini-Mental State Examination (MMSE). We calculate the average of MMSE scores per individual~\footnote{We don't calculate longitudinal changes in the behavioural measures due to 
missing values in the datasets. } and the average difference in the communication marker between the same individual's adjacent sessions. Positive values of change indicate improvement in communication over time while negative values denote the opposite. 
Similarly, high MMSE scores are indicative of better cognitive function (refer to Appx.~\ref{app:biomarkers} for details on MMSE). Figure~\ref{fig:mmse} illustrates the correlation between averaged longitudinal changes in the communication marker and average MMSE scores. We notice that people with a high MMSE score either improve or exhibit minor changes in communication over time. On the other hand, the communication marker decreases for those people with low MMSE scores. Overall, we found a Pearson correlation of 0.61 ($p=4.48e^{-8}$) between changes in MMSE and the average difference in the communication marker over time.

\begin{figure}[hbt!]
\centering
\includegraphics[width=.50\textwidth]{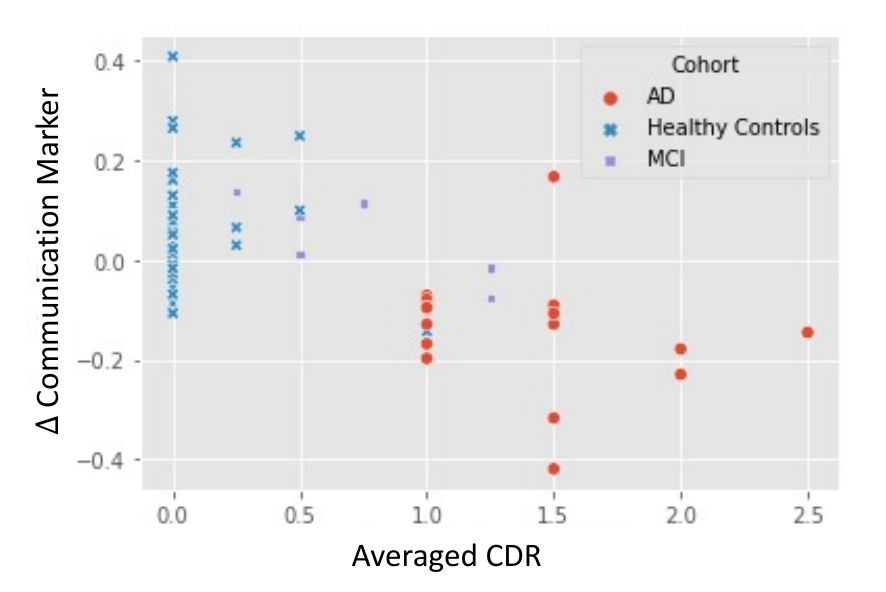}
\caption{ Association between average longitudinal changes in  communication marker and average values of the Clinical Dementia Rating (CDR), across 
cohorts.} 
\label{fig:cdr}
\end{figure}

Similarly, we investigate the association between average longitudinal changes in the communication marker with the Clinical Dementia Rating (CDR). Here, the higher the CDR, the lower the cognitive function (see Appx.~\ref{app:biomarkers} for details on CDR). Figure~\ref{fig:cdr} illustrates the association between average longitudinal changes in the communication marker and CDR. We note that people with low average values of CDR (i.e., CDR $\in [0,1)$) improved their communication over time. This is presumably because subjects are able to remember and do better at the CTP description task when seeing it again~\cite{goldberg2015practice}. However, people with moderate to high levels of CDR (i.e., CDR $\in [1,3]$) exhibit impairment in communication over time. Overall, we found a Pearson correlation of 0.56 ($p=6.67e^{-7}$) between average CDR values and average values in changes for the communication over time.

We observe that people with AD with severe cognitive impairment, i.e. MMSE ranging from 14-18 and CDR from 2-2.5, did not exhibit a severe decrease in the  communication marker over time. We attribute this to a ceiling effect. Indeed, a meta-analysis shows that the communication marker for people with the lowest behaviour scores was already much lower at the onset compared to those AD participants with higher behaviour scores. 

\subsection{Linguistic Disorder Markers}
\label{sec:disorder_markers}
We investigate how different linguistic disorder markers capture the impact on individuals' speech. We compute the markers as the percentage of occurrence of each of the language disorders in Table~\ref{tbl:lan_dis_type} using the normalized probability estimates of the model (for details on how the markers are obtained, see Sec.~\ref{sec:task}). 
Table~\ref{tbl:disorder_markers} provides the average percentage value of each linguistic disorder marker per cohort as well as corresponding percent changes from the end to the start of the study. 
The higher the percentage of a marker the more prevalent the language disorder.

\begin{table}[h!]

\centering 

\resizebox{\columnwidth}{!}{%
\begin{tabular}{llcccccc}
 \toprule
 \multicolumn{1}{c}{} &
 \multicolumn{2}{c}{\textbf{Anomia}} &
 \multicolumn{2}{c}{\textbf{Disfluency}} &
 \multicolumn{2}{c}{\textbf{Agrammatism}}\\
\cmidrule(r){2-3} \cmidrule(r){4-5} \cmidrule(r){6-7}

\textbf{Cohort} & \textbf{Marker}  & \textbf{$\Delta$}  & \textbf{Marker}  & \textbf{$\Delta$} & \textbf{Marker}  & \textbf{$\Delta$}     \\
 \hline
 Healthy & 1.11 & +1.12  & 15.43  &  +1.91 & 11.41 & -3.85 \\ \hline
 MCI & 1.94  & +1.35  & 21.66  & -4.75  & 13.31 & -0.44 \\ \hline
AD & \textbf{5.58}  & +2.18  & \textbf{25.11}  & +8.82  & 15.86 & \textbf{+8.95} \\ 
 \bottomrule
\end{tabular}
}
\caption{Percentage of 
language disorders 
as captured by the corresponding linguistic markers across 
cohorts. Marker: Average of marker within a cohort. $\Delta_{(end-start)}$: Average change of the marker from the end to the beginning of the study. Negative numbers imply improvement over time. Numbers in bold denote significant difference across cohorts.}
\label{tbl:disorder_markers}
\end{table}

We note that people across all cohorts exhibit disfluency. However, the disfluency marker was significantly higher for people with AD compared to healthy controls ($p < 0.05$).$^{\ref{test}}$ The MCI cohort exhibits improvement in disfluency over the study ($\Delta$=-4.75\% in Table~\ref{tbl:disorder_markers}). 
Anomia is characteristic of people with AD~\cite{botha2019primary} and despite being less prevalent overall is significantly$^{~\ref{test}}$ higher for the AD cohort.  
Although agrammatism is more prominent in people with AD, there is no significant difference across cohorts. We attribute this to the same relative ratio of aggramatism in healthy controls and people with dementia in the training data (see Table~\ref{tbl:data_train} where Sub:Aggr$\approx$.55 in both cases) rather than the sensitivity of the marker itself. 
Indeed the aggramatism marker captures that people with AD exhibit a significant change in syntactic disturbance over time (+8.95\% in the value of the marker) whereas the rest of the cohorts improved over time.

Overall, the linguistic disorder markers were effective in screening and monitoring AD where gradual language impairment ensues.

\section{Conclusion}
We are the first to introduce reformulated NLP tasks for learning language disorder patterns from transcribed speech in dementia datasets by forcing a  pre-trained language model to obtain signal from the language itself. 
Our experiments show that NLP tasks encapsulating contextual information and enhancing the gradient signal with linguistic patterns benefit performance. We use the probability estimates of the model with highest macro-F1 to construct digital markers measuring communication ability and the occurrence of various language disorders in the speech of people with dementia and healthy controls.  Longitudinal analysis shows that the digital communication marker is able to assess the quality of communication and distinguish between people with MCI, Alzheimer’s Disease (AD) and healthy controls. A comparison against existing linguistic approaches for capturing language impairment shows the superiority of our proposed communication marker. Moreover, the latter correlates significantly with two widely used clinical behaviour markers.  
Finally, our proposed linguistic disorder markers prove effective for screening and monitoring AD and provide useful insights into longitudinal change in linguistic ability. In the future we will explore large pre-trained generative transformers and automatic generation of templates to improve performance on capturing linguistic disorder patterns.
\section*{Limitations}

Monitoring dementia using computational linguistics approaches is an important topic. Previous work has primarily focused on learning language discrimination across healthy controls and people with AD, ignoring longitudinal language disorders. In this work, we use DementiaBank to capture longitudinal linguistic disorder patterns that characterize people living with dementia. Currently, DementiaBank is the largest available longitudinal dementia dataset. A limitation of DementiaBank is that the longitudinal aspect is limited, spanning up to 5 sessions/descriptions maximum per individual, with most participants contributing up to two narratives. Moreover, the number of participants is relatively small, especially for the mild-cognitive impairment (MCI) cohort. Finally, descriptions are elicited through the Cookie Theft Picture (CTP), ignoring interactive aspects of
everyday conversational interaction. The Carolinas Conversation Collection dataset~\cite{pope2011finding} contains more natural conversations between patients and clinical practitioners. However, it  only contains speech data from people with AD and no equivalent data for healthy controls. In the future, we aim to address these limitations by investigating the generalisability of our proposed digital language disorder markers on a novel fine-grained longitudinal multi-modal dataset from people with dementia over several months in a natural setting (currently under review).

In this study, we used manually transcribed data from DementiaBank and  its paralinguistic information to annotate transcribed turns. In a real-world scenario, participants mostly provide speech via a speech elicitation task. This implies that the introduced method requires an automatic speech recognition (ASR) system robust to various sources of noise to be operationalized. ASR for mental health is currently underexplored, with most transcription work being done by humans.

It may be that the proposed digital linguistic markers become a less accurate means for monitoring dementia when people experience other comorbidities, neurodegenerative and mental illnesses, that significantly affect speech and language. Indeed, cognitive-linguistic function is a strong biomarker for neuropsychological health \cite{voleti2019review}.

Finally, there is a great deal of variability to be expected in speech and language data affecting the sensitivity of the proposed digital linguistic markers. Both speech and language are impacted by speaker identity, context, background noise, spoken language etc. Moreover, people may vary in their use of language due to various social contexts and conditions, a.k.a., style-shifting \cite{coupland2007style}. Both inter and intra-speaker variability in language could affect the sensitivity of the proposed digital markers. While it is possible to tackle intra-speaker language variability, e.g., by integrating speaker-dependent information to the language, the inter-speaker variability remains an open-challenging research question.
\section*{Ethics Statement}

Our work does not involve ethical considerations around the analysis of the DementiaBank and ADReSS corpora as they are widely used. For DementiaBank, ethics was obtained by the original research team by James Backer and participating individuals consented to share their data following a larger protocol administered by the Alzheimer and Related Dementias Study at the University of Pittsburgh School of Medicine \cite{becker1994natural}. Access to the data is password protected and restricted to those signing an agreement. For ADReSS, ethics was obtained by the original research team by Brian MacWhinney that collected the data for ADReSS challenge. Access to the data requires membership of DementiaBank and a non-disclosure agreement between the stakeholders and the research team.  

This work uses transcribed dementia data to identify changes in cognitive status considering individuals’ language disorders. Research  Potential risks from the application of our work in being able to identify cognitive decline in individuals are akin to those who misuse personal information for their own profit without considering the impact and the social consequences in the broader community. Potential mitigation strategies include running the software on authorised servers, with encrypted data during transfer, and anonymization of data prior to analysis. Another possibility would be to perform on-device processing (e.g. on individuals’ computers or other devices) for identifying changes in cognition and the results of the analysis would only be shared with authorised individuals. Individuals would be consented before any of our software would be run on their data.

\section*{Acknowledgements}
This work was supported by a UKRI/EPSRC
Turing AI Fellowship to Maria Liakata (grant
EP/V030302/1), the Alan Turing Institute (grant
EP/N510129/1), and Wellcome Trust MEDEA (grant 213939). Matthew Purver acknowledges financial support from the UK EPSRC via the projects Sodestream (EP/S033564/1) and ARCIDUCA (EP/W001632/1), and from the Slovenian Research Agency grant for research core funding P2-0103.


\bibliography{acl_latex}
\bibliographystyle{acl_natbib}

\appendix
\appendix

\section{The Cookie Theft Picture}
\label{app:ctp}

\begin{figure}[ht]
\centering
\includegraphics[width=.5\textwidth]{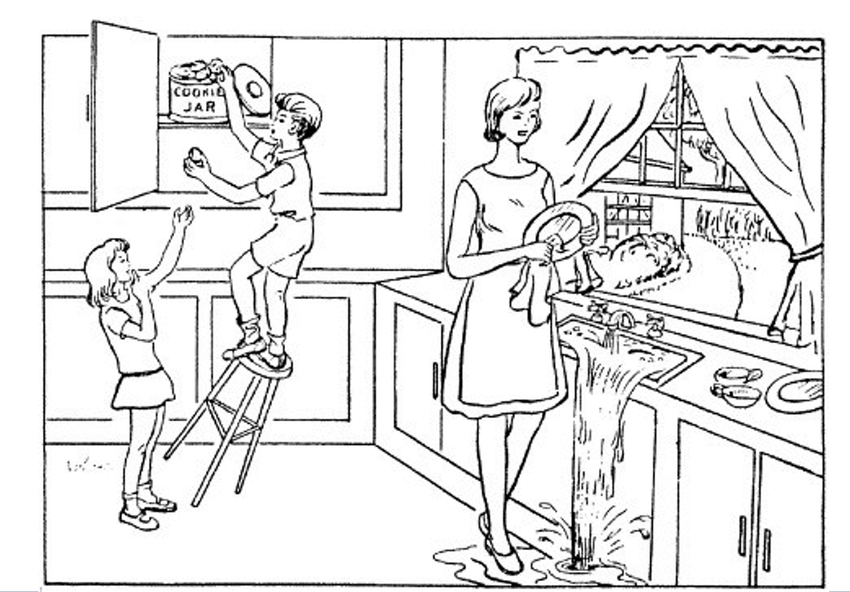}
\caption{The Cookie Theft Picture from the Boston Diagnostic Aphasia Examination~\cite{ctp}.}
\label{fig:ctp}
\end{figure}

For the PD task, the examiner asks subjects to describe the picture (see Fig. \ref{fig:ctp}) by saying, "Tell me everything you see going on in this picture". Then subjects might say, "there is a mother who is drying dishes next to the sink in the kitchen. She is not paying attention and has left the tap on. As a result, water overflows from the sink. Meanwhile, two children attempt to make cookies from a jar when their mother is not looking. One of the children, a boy, has climbed onto a stool to get up to the cupboard where the cookie jar is stored. The stool is rocking precariously. The other child, a girl, is standing next to the stool and has her hand outstretched ready to be given cookies.

\section{Dementia datasets}
\label{app:pitt}

\subsection{DentiaBank}
The dataset was gathered longitudinally between 1983 and 1988 as part of the Alzheimer Research Program at the University of Pittsburgh. The study initially enrolled 319 participants according to the following eligibility criteria: all the participants were required to be above 44 years old, have at least seven years of education, have no history of major nervous system disorders, and have an initial Mini-Mental State Examination score above 10. Finally, the cohort consisted of 282 subjects.  In particular, the cohort included 101 healthy control subjects (HC) and 181 Alzheimer’s disease subjects (AD). An extensive neuropsychological assessment was conducted on the participants, including verbal tasks and the Mini-Mental State Examination (MMSE). 

\subsection{ADReSS}
ADReSS is a benchmark dataset of spontaneous speech, which is acoustically pre-processed and balanced in terms of age and gender. The dataset entails transcribed speech of 78 non-AD subjects and 78 AD subjects of 35 males and 43 females for each of the cohorts. The dataset was made available for the ADReSS challenge consisted of two tasks: a) an AD classification task, where the task required one to produce a model to predict the label (AD or non-AD) for a speech session and b) an MMSE score regression task, where the task required one to create a model to infer the subject's Mini-Mental Status Examination (MMSE) score based on speech and/or language data.

\section{Coding Scheme for the Annotation of Transcribed Utterances.}
\label{app:chat}

Table~\ref{tbl:chat} lists the codes we used to annotate transcribed speech utterances in accordance with the CHAT protocol~\cite{macwhinney2017tools}. Moreover, we used the code $[+exc]$ to filter out non-descriptive utterances from the Cookie Thief Picture (CTP) description task (e.g., "Yeah that's it."). As shown in Table~\ref{tbl:chat}, the manifestation granularity varies across different language disorders. For example, anomia is exhibited through various symptoms in language.

\begin{table}[htbp]
\centering 

\resizebox{\columnwidth}{!}{%
\begin{tabular}{lcp{5cm}}
 \toprule
 
\textbf{Disorder} &  \textbf{Code} & \textbf{Manifestation in Language}   \\
 \hline
 Agrammatism & $[+ gram]$  & Agrammatic and paragrammatic speech.\\ \hline
 Disfluency & $[/]$  & Word or phrase repetition. \\
  & $[//]$  & Word or phrase revision. \\
 & $\&+$  & Phonological fragment. \\ \hline
 Anomia & $+ es$  & Empty speech. \\
  & $+...$  & Termination of an incomplete utterance.\\
 & $[+ cir]$  & Talking around words/concepts. \\ 
 & $[+ jar]$  & Fluent and prosodically correct but largely meaningless speech. \\ 
 \hline
 Disruptive & $[+ exc]$  & Non-descriptive speech. \\

 \bottomrule
\end{tabular}
}
\caption{Coding scheme used for the annotation of transcribed speech utterances following the CHAT protocol~\cite{macwhinney2017tools}. } 
\label{tbl:chat}
\end{table}

\section{Experimental Settings}
\label{app:train}

We used a grid search optimization technique to optimize the parameters. For consistency, we used the same experimental settings for all models. We first fine-tuned all models by performing a twenty-times grid search over their parameter pool. We empirically experimented with learning rate ($lr$): $lr \in \{0.00001,0.00002,0.00005,0.0001,0.0002\}$, batch size ($bs$): $bs \in \{16,32,64,128\}$ and optimization ($O$): $O \in \{AdamW,Adam\}$. After the fine-tuning process, we trained again all the models for 50 epochs with 4 epochs early stopping, three times. We reported the average performance on the test set for all experiments. Model checkpoints were selected based on the minimum validation loss. Experiments were conducted on two GPUs, Nvidia V-100.

For fine-tuning RoBERTa with MLM jointly, we suggest the weights (1/0.5139) for the classification objective and (1/2.4149) for the MLM objective.

To investigate how various language disorders involve with the progression of dementia, we construct anomia, disfluency, and agrammatism markers, by first extracting the corresponding model's probability estimates for each utterance, i.e., $p(y_i^\mathcal{L} \; | \;  y_{i}=y_{i}^*)$, where $y_{i}^*$ $\in \{anomia, disfluency, agrammatism\}$. We then obtain averaged probabilities over the entire session (description of the CTP).

\section{Clinical Behavioural Markers}
\label{app:biomarkers}
\subsection{ Mini-Mental State Examination (MMSE)}

The Mini-Mental State Examination (MMSE) has been the most common method for diagnosing AD and other neurodegenerative diseases affecting the brain. It was devised in 1975 by Folstein et al. as a simple standardized test for evaluating the cognitive performance of subjects, and where appropriate to qualify and quantify their deficit. It is now the standard bearer for the neuropsychological evaluation of dementia, mild cognitive impairment, and AD.

The MMSE was designed to give a practical clinical assessment of change in cognitive status in geriatric patients. It covers the person’s orientation to time and place, recall ability, short-term memory, and arithmetic ability. It may be used as a screening test for cognitive loss or as a brief bedside cognitive assessment. By definition, it cannot be used to diagnose dementia, yet this has turned into its main purpose.

The MMSE includes 11 items, divided into 2 sections. The first requires verbal responses to orientation, memory, and attention questions. The second section requires reading and writing and covers ability to name, follow verbal and written commands, write a sentence, and copy a polygon. All questions are asked in a specific order and can be scored immediately by summing the points assigned to each successfully completed task; the maximum score is 30. A score of 25 or higher is classed as normal. If the score is below 24, the result is usually considered to be abnormal, indicating possible cognitive impairment. The MMSE has been found to be sensitive to the severity of dementia in patients with Alzheimer’s disease (AD). The total score is useful in documenting cognitive change over time.

\subsection{Clinical Dementia Rating (CDR)}
The Clinical Dementia Rating (CDR) is a global rating device that was first introduced in a prospective study of patients with mild “senile dementia of AD type” (SDAT) in 1982 (Hughes et al., 1982). New and revised CDR scoring rules were later introduced (Berg, 1988; Morris, 1993; Morris et al., 1997). CDR is estimated on the basis of a semistructured interview of the subject and the caregiver (informant) and on the clinical judgment of the clinician. CDR is calculated on the basis of testing six different cognitive and behavioral domains such as memory, orientation, judgment and problem solving, community affairs, home and hobbies performance, and personal care. The CDR is based on a scale of 0–3: no dementia (CDR = 0), questionable dementia (CDR = 0.5), MCI (CDR = 1), moderate cognitive impairment (CDR = 2), and severe cognitive impairment (CDR = 3). Two sets of questions are asked, one for the informant and another for the subject. The set for the informant includes questions about the subject’s memory problem, judgment and problem solving ability of the subject, community affairs of the subject, home life and hobbies of the subject, and personal questions related to the subject. The set for subject includes memory-related questions, orientation-related questions, and questions about judgment and problem-solving ability.

\end{document}